\documentclass[conference]{IEEEtran}
\IEEEoverridecommandlockouts
\usepackage{algorithmic}
\usepackage{cite}
\usepackage{amsmath,amssymb,amsfonts}
\usepackage{bm}
\usepackage{dsfont}
\usepackage{booktabs}
\usepackage{multirow}
\usepackage{placeins}
\usepackage{graphicx}
\usepackage{textcomp}
\usepackage{xcolor}
\usepackage{url}
\usepackage{hyperref}
\def\BibTeX{{\rm B\kern-.05em{\sc i\kern-.025em b}\kern-.08em
    T\kern-.1667em\lower.7ex\hbox{E}\kern-.125emX}}

\begin{document}

\title{SegMate: Asymmetric Attention-Based Lightweight Architecture for Efficient Multi-Organ Segmentation}

\author{\IEEEauthorblockN{Andrei-Alexandru Bunea}
\IEEEauthorblockA{\textit{POLITEHNICA Bucharest} \\
Bucharest, Romania\\
\texttt{andrei.bunea@upb.ro}
}
\and
\IEEEauthorblockN{Dan-Matei Popovici}
\IEEEauthorblockA{\textit{POLITEHNICA Bucharest} \\
Bucharest, Romania\\
\texttt{pdmatei@gmail.com}
}
\and
\IEEEauthorblockN{Radu Tudor Ionescu}
\IEEEauthorblockA{\textit{University of Bucharest} \\
Bucharest, Romania\\
\texttt{raducu.ionescu@gmail.com}
}
}
\maketitle

\begin{abstract}
State-of-the-art models for medical image segmentation achieve excellent accuracy, but require substantial computational resources, limiting deployment in typical resource-constrained clinical settings. We present SegMate, an efficient 2.5D framework that achieves state-of-the-art accuracy, while considerably reducing computational requirements. Our efficient design is the result of meticulously integrating asymmetric architectures, attention mechanisms, multi-scale feature fusion, slice-based positional conditioning, and multi-task optimization.
We demonstrate the efficiency-accuracy trade-off of our framework across three modern backbones (EfficientNetV2-M, MambaOut-Tiny, FastViT-T12). We perform experiments on three datasets: TotalSegmentator, SegTHOR and AMOS22. Compared with the vanilla models, SegMate reduces computation (GFLOPs) by up to $\mathbf{2.5\times}$ and memory footprint (VRAM) by up to $\mathbf{2.1\times}$, while generally registering performance gains of around 1\%. On TotalSegmentator, we achieve a Dice score of 93.51\% with only 295 MB peak GPU memory. Zero-shot cross-dataset evaluations on SegTHOR and AMOS22 demonstrate strong generalization, with Dice scores of up to 86.85\% and 89.35\%, respectively. SegMate variants in this study empower single GPU setups to process up to ${\sim}$10,000 volumes per hour with just 74 J per volume using batch inference, enabling population-scale CT processing on commodity hardware. We release our open-source code at \url{https://github.com/andreibunea99/SegMate}.
\end{abstract}


\section{Introduction}

Radiation therapy represents a fundamental method in treating cancer, used in approximately 50\% of all treatment plans for cancer patients \cite{delaney2005role}.
Precision is critical in the delineation process of target volumes and organs-at-risk (OARs), as an accurate segmentation enables an optimal dose distributed to the targeted tumors, while minimizing the exposure to healthy tissues \cite{brouwer2015ct}. Manual segmentation is the gold standard in clinical practice, but the task requires maximal attention, lasting 30-60 minutes per patient \cite{sharp2014vision}. Automated multi-organ segmentation can reduce delineation time from 30-60 minutes to just seconds, cutting resource costs by up to 95\%, while also capturing subtle features that are difficult to identify under clinical time pressure.

While state-of-the-art segmentation models achieve high accuracy, they usually require increased computational resources (8–16GB GPU memory) during training and inference~\cite{ma2022flare}. Although a single study can run on modest hardware given sufficient time, processing at scale uncovers the real constraint.
Imaging archives grow by thousands of CT volumes per institution every year, and applications such as opportunistic screening, cohort-level analysis, and large-scale dataset curation require segmentation on the entire archive, rather than individual studies. At this scale, the per-volume compute, memory, and energy costs determine whether processing the entire archive takes hours on a single workstation or days on a data-center cluster, turning efficient segmentation into a big data processing problem, on top of the clinical deployment constraint.

\begin{figure*}[t]
    \centering
    \includegraphics[width=0.78\linewidth]{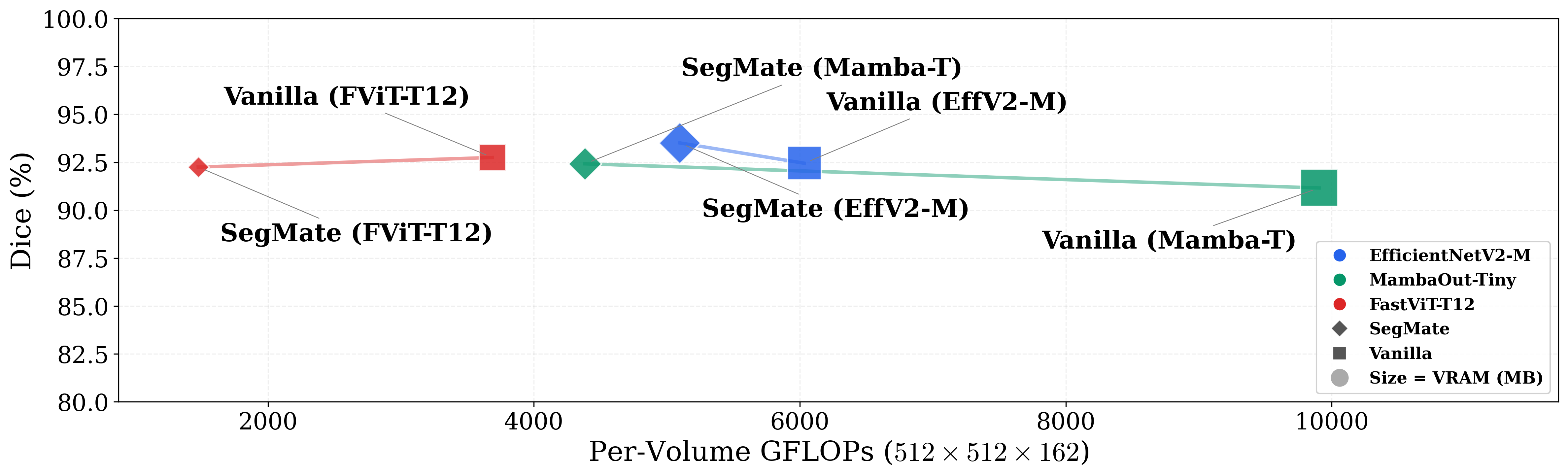}
    \vspace{-0.3cm}
    \caption{Segmentation accuracy vs.~GFLOPs per volume ($\approx$162 slices) on TotalSegmentator. Point size indicates VRAM usage. All SegMate variants are below 300 MB. SegMate preserves segmentation performance (Dice), while significantly reducing GFLOPs. Best viewed in color.}
    \label{fig:tradeoff}
\end{figure*}

To bridge the accessibility gap, we propose SegMate, a novel framework that preserves segmentation performance, while significantly reducing compute time and memory footprint (see Fig.~\ref{fig:tradeoff}). Our efficiency-oriented design is based on a meticulous integration of multiple components into a joint framework: (1) early slice fusion via channel attention, (2) asymmetric auto-encoder architecture with a lighter decoder, (3) dual attention mechanisms, (4) multi-scale feature fusion, (5) slice-based positional embeddings, and (6) multi-task optimization.

We carry out experiments on three datasets (TotalSegmentator \cite{wasserthal2023totalsegmentator}, SegTHOR \cite{lambert2020segthor} and AMOS22 \cite{ji2022amos}) to demonstrate that SegMate can be integrated into various architectures (EfficientNetV2-M 
\cite{tan2021efficientnetv2}, MambaOut-Tiny \cite{yu2024mambaout}, FastViT-T12 \cite{vasu2023fastvit}) to improve efficiency (in terms of speed and memory), while preserving accuracy. We also show that SegMate leads to improved generalization capacity in zero-shot evaluation setups. Moreover, we perform ablation studies to justify the integration of the proposed components into the joint framework, considering both efficiency and effectiveness.

In summary, our contribution is twofold:
\begin{itemize}
    \item We introduce SegMate, a novel framework that can be integrated into multiple medical image segmentation models to boost efficiency, while preserving effectiveness. 
    \item To demonstrate the wide applicability of SegMate, we perform comprehensive experiments across multiple datasets, evaluation scenarios, and neural architectures.
\end{itemize}

\section{Related Work}

Machine learning solutions for segmentation in medical images have significantly evolved over the past decades with advances in deep learning architectures and transformers \cite{litjens2017survey}. The introduction of U-Net \cite{ronneberger2015unet} marked a cornerstone in precise delineation, leveraging the auto-encoder architecture to capture features at various scales in images. 
Many best-performing models build on enhanced U-Net variants, such as U-Net++ \cite{zhou2018unetpp}, Attention U-Net \cite{oktay2018attention}, and 3D U-Net \cite{cicek20163d}. nnU-Net \cite{isensee2021nnu} demonstrates that extensive data preprocessing could drive the performance even higher, enabling an automated hyperparameter optimization mechanism on the training dataset. However, this approach requires substantial computational resources and complex setup procedures. In the last few years, transformer-based architectures like Swin-UNETR \cite{hatamizadeh2022swin}, TransUNet \cite{chen2021transunet}, and UNETR \cite{hatamizadeh2022unetr} pushed accuracy boundaries (94-95\% Dice) at the cost of high GPU memory requirements.

\subsection{Foundation models and volumetric adaptation}
VISTA3D \cite{he2025vista3d} establishes a unified 3D segmentation foundation model trained on large-scale annotated CT data. Bio2Vol \cite{zhuang2025bio2vol} adapts BiomedParse, a 2D foundation model, to volumetric segmentation, achieving 85.45\% Dice across 15 AMOS22 organs, though at substantially higher computational cost than task-specific models. The Post-Axial Refiner (PaR) \cite{huang2025par} takes a different angle: it disentangles axial-plane features through a plug-and-play module appended after decoding. 
Notably, SegMate achieves per-organ results competitive with these methods, without foundation-model-scale pre-training or 3D convolutions, at a fraction of the GPU memory.

\subsection{Efficient architectures in medical imaging} 
While scaling deep neural networks by adding hidden layers and increasing parameter counts is the main research focus nowadays, some studies concentrate on creating efficient architectures that can demonstrate equivalent performance to heavy models, but with limited computing and memory resources. Generic architectures, such as EfficientNet \cite{tan2019efficientnet} and MobileNet \cite{howard2017mobilenets}, have been employed in medical image analysis in previous work \cite{kus2024medsegbench}. Some recent studies target the efficiency bottleneck from the decoder side. EffiDec3D \cite{rahman2025effidec3d} introduces Base Channel Scaling and High-Resolution Removal to cut decoder parameters by up to 96\%. EffiDec3D is applied as a model-agnostic plugin to SwinUNETR and 3D UX-Net. Its successor, EfficientMedNeXt \cite{rahman2025efficientmednext}, pairs these decoder optimizations with Dilated Multi-Receptive Field Blocks that capture multi-scale context through parallel dilated convolutions. 
Both works optimize \emph{decoder channels}. In contrast, SegMate takes a complementary route by optimizing \emph{attention placement} and encoder-decoder asymmetry, keeping the full decoder resolution intact, while reducing activation memory through lightweight Squeeze-and-Excitation (SE) \cite{hu2018squeeze} gating.

\section{Method}


\subsection{Architecture overview}

\begin{figure*}[t]
    \centering
    \includegraphics[width=0.78\linewidth]{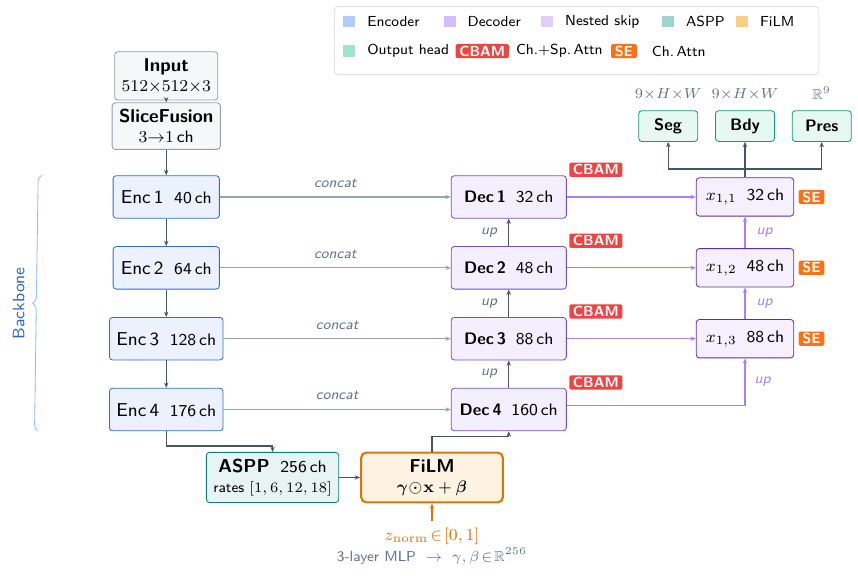}
    \vspace{-0.3cm}
    \caption{Our SegMate architecture. CBAM modules are positioned inside the main decoder blocks after encoder-decoder fusion. SE modules gate nested skip blocks that cascade cross-scale features. FiLM conditions the bottleneck on normalized slice position $z_{\mathrm{norm}}$. Best viewed in color.}
    \label{fig:architecture}
\end{figure*}

SegMate is a novel framework that is meticulously designed to harmoniously integrate multiple architectural changes, aiming to boost the computational speed and reduce the memory footprint of auto-encoders applied in medical image segmentation. Our changes are carefully crafted to make every processing component more efficient, starting from the input and going all the way to the output. For the input, we introduce SliceFusion, an attention mechanism that fuses the three input slices into a single slice, going from 2.5D to 2D processing. For the architecture, we introduce several changes: (1) an asymmetric encoder-decoder design with a lighter decoder, (2) Squeeze-and-Excitation (SE) \cite{hu2018squeeze} blocks in dense decoder skip connections for inter-level feature fusion, (3) Convolutional Block Attention Module (CBAM) \cite{woo2018cbam} in the main decoder path for encoder-decoder integration, (4) slice-positional encoding via Feature-wise Linear Modulation (FiLM) \cite{perez2018film}. For the output, we attach three prediction heads that produce segmentation masks, organ boundaries, and organ presence maps. The overall architecture is illustrated in Fig.~\ref{fig:architecture}.



\subsection{2.5D to 2D}
While 3D networks provide full volumetric context, they impose prohibitive memory costs. 2.5D approaches process stacks of adjacent slices, offering a middle ground between pure 2D efficiency and 3D spatial awareness \cite{han2021spine}. We confer vertical (axial) spatial information to our architecture using a 2.5D approach. However, we apply SliceFusion, an attention mechanism that fuses the three input slices into a single slice, thus going from 2.5D to 2D processing for the entire architecture. For each target slice $t$, we stack three adjacent axial slices $(t{-}1, t, t{+}1)$ as an input tensor. Our learned fusion module (SliceFusion)  
comprises a 2D conv layer with 16 filters of $3{\times}3$, a batch normalization layer, a SiLU activation layer, and another 2D conv layer with one filter of $1{\times}1$. SliceFusion keeps the processing to a minimum, while still harnessing 2.5D information.



\subsection{Slice positional conditioning}
Since our architecture does not process entire volumes, segmentation performance can be affected. To leverage spatial coherence across the Z-axis without explicit 3D convolutions, we propose to condition the model on the normalized slice position $z_{\text{norm}} \in [0,1]$, computed as the relative position of each slice within the patient-specific volume extent. For this step, employ a generic conditioning module called Feature-wise Linear Modulation (FiLM) \cite{perez2018film}. After the ASPP bottleneck \cite{ding2023real}, we apply a learned affine transformations conditioned on slice position:
\begin{equation}
    \mathbf{y} = \gamma(z_{\text{norm}}) \odot \mathbf{x} + \beta(z_{\text{norm}}),
\end{equation}
where $\mathbf{x} \in \mathbb{R}^{H \times W \times C}$ are the bottleneck features, $\gamma, \beta: [0,1] \rightarrow \mathbb{R}^{C}$ are learned functions implemented as three-layer MLPs with hidden dimension 128, and $\odot$ denotes element-wise multiplication. This modulation allows the network to implicitly learn organ appearance variations across different anatomical levels (e.g.~upper vs.~lower thorax) without increasing inference memory or requiring 3D convolutions. Note that $z_{\text{norm}}$ is defined relative to the acquired axial extent of each scan, rather than to a body-part coordinate system.

\subsection{Synergic dual-attention design}
Attention-based segmentation models generally apply a single attention mechanism uniformly throughout the network \cite{Georgescu-WACV-2023}. Instead, we propose a dual-attention design, assigning distinct roles to SE and CBAM blocks. Our nested decoder implements dense skip connections between adjacent decoder levels, similar to a pruned UNet++. At each nested node, SE blocks \cite{hu2018squeeze} perform channel-wise recalibration: global average pooling compresses spatial information into a channel descriptor, which is mapped by a two-layer bottleneck (with $16{\times}$ reduction) to per-channel gates. This lightweight gating selects informative features during iterative cross-scale fusion from adjacent decoder levels without significant computational overhead. In the main decoder path, encoder features are concatenated with up-sampled decoder features and passed through CBAM \cite{woo2018cbam}. Beyond channel recalibration, CBAM adds a spatial attention map that localizes informative regions via average and max pooled projections followed by a $7{\times}7$ convolution. This dual gating with channel and spatial attention recovers fine-grained details lost during downsampling in the encoder path, especially for small organs, e.g.~esophagus, trachea, etc.

\subsection{Asymmetric architecture}
Inspired by masked auto-encoders \cite{He-CVPR-2022}, we employ an asymmetric design combining a high-capacity encoder paired with a compact decoder (at most 160 channels) via an ASPP bottleneck, allowing SegMate to contain 86\% more parameters than the standard U-Net, yet consume less GPU memory. The paradox comes from the SE and CBAM blocks that act as lightweight gating mechanisms, keeping decoder activations small, with a maximum decoder width of 160 channels.

\begin{table*}[t]
\caption{Dice scores (\%) over all organs, across three datasets. For SegTHOR and AMOS22, we show zero-shot (ZS) and fine-tuning (FT) transfer from TotalSegmentator. Per-volume GFLOPs for a 512×512×162 CT volume with 162 sequential slices. Best per-backbone is in \textbf{bold}. Scores represent averages over 3 runs.}
\label{tab:unified_overview}
\vspace{-0.3cm}
\setlength\tabcolsep{0.11cm}
\begin{center}
{%
\begin{tabular}{|l|l|c|cc|cc|c|c|}
\hline
\textbf{Backbone} & \textbf{Method} & \textbf{TotalSeg} & \textbf{SegTHOR ZS} & \textbf{SegTHOR FT} & \textbf{AMOS22 ZS} & \textbf{AMOS22 FT} & \textbf{VRAM} & \textbf{GFLOPs} \\
\hline
\multirow{2}{*}{EfficientNetV2-M}
  & SegMate  & \textbf{93.51} & \textbf{86.85} & \textbf{88.51} & \textbf{89.35} & \textbf{91.29} & \textbf{295 MB} & \textbf{5,097} \\
  & Vanilla & 92.45 & 85.36 & 88.06 & 89.29 & 91.22 & 374 MB & 6,034\\
\hline
\multirow{2}{*}{MambaOut-T}
  & SegMate  & \textbf{92.42} & \textbf{86.46} & \textbf{86.67} & \textbf{89.14} & \textbf{90.44} & \textbf{208 MB} & \textbf{4,384} \\
  & Vanilla & 91.16 & 83.11 & 86.66 & 88.22 & 90.28 & 431 MB & 9,905\\
\hline
\multirow{2}{*}{FastViT-T12}
  & SegMate  & 92.25 & \textbf{84.23} & \textbf{87.13} & \textbf{88.87} & \textbf{91.52} & \textbf{120 MB} & \textbf{1,476} \\
  & Vanilla & \textbf{92.75} & 83.19 & 86.02 & 88.24 & 88.51 & 253 MB & 3,688\\
\hline
\end{tabular}%
}
\end{center}
\end{table*}

\subsection{Multi-task boundary enhancement}
SegMate jointly predicts segmentation masks, organ boundaries, and organ presence using three prediction heads. The total loss is determined using the weighted sum of the losses corresponding to each head:
\begin{equation}
    \mathcal{L}_{\text{total}} = \lambda_{\text{seg}} \cdot \mathcal{L}_{\text{seg}} + \lambda_{\text{bdy}} \cdot \mathcal{L}_{\text{bdy}} + \lambda_{\text{prs}} \cdot\mathcal{L}_{\text{prs}}.
\end{equation}
The \textbf{segmentation head} combines Dice, focal, and cross-entropy losses:
\begin{equation}
    \mathcal{L}_{\text{seg}} = \lambda_{\text{Dice}} \cdot \left( \alpha \cdot \mathcal{L}_{\text{Dice}} + (1{-}\alpha) \cdot\mathcal{L}_{\text{focal}} \right) + \lambda_{\text{CE}} \cdot\mathcal{L}_{\text{CE}},
\end{equation}
where $\mathcal{L}_{\text{Dice}}$ mitigates class imbalance, $\mathcal{L}_{\text{focal}}$ emphasizes hard pixels, and $\mathcal{L}_{\text{CE}}$ stabilizes training. The \textbf{boundary head} predicts edge probability maps via a $1{\times}1$ projection, supervised by Dice loss against Sobel-extracted ground-truth edges. The \textbf{presence head} uses global average pooling followed by a fully connected layer to produce a vector $\hat{p} \in[0,1]^K$, trained with binary cross-entropy against $y_k{=}\mathbf{1}\!\left(\sum Y_{\text{true}}^{(k)}{>}0\right)$, suppressing false positives on slices where an organ is absent. Here, $K$ is the number of organs.


\subsection{Design rationale}
The contribution of SegMate lies in the design and motivation of the architecture which combines and repurposes existing components (SE, CBAM, FiLM, ASPP) in a novel perspective to produce state-of-the-art accuracy for resource constrained clinical settings. We assign the cheapest gating mechanism (channel-only produced by SE blocks) to the numerous nested skip nodes, where a lightweight recalibration is performed repeatedly across different scales. The CBAM blocks are heavier, having both channel and spatial gating, and we reserve these for the main decoder path, where the encoder features must be spatially re-localized after fusion. The efficiency of the architecture also comes from the asymmetry of the design, as SegMate combines a high-capacity encoder to learn complex structures with a narrow decoder (under 160 channels) that keeps the high-resolution activations small. To address the axial context gap between 2.5D and 3D architectures, SegMate introduces a conditioning module as a cheap yet effective method to keep the learning aware of the general overview of the CT volume. Each design choice trades a small cost in performance, for a considerable efficiency gain, as shown in Figure \ref{fig:tradeoff}.

\section{Experiments and Results}

\subsection{Datasets and experimental setup}
\noindent\textbf{Datasets.}
We perform experiments on TotalSegmentator v2.0.1 \cite{wasserthal2023totalsegmentator}, selecting a subset of 9 organs spanning a wide range of sizes, representative for general assumptions in OAR delineation tasks: left/right lung, spinal cord, esophagus, liver, left/right kidney, aorta and trachea. These structures cover the full spectrum of feature sizes, from the large lungs and liver to the small esophagus, trachea and spinal cord, and were primarily dictated by the label overlap with SegTHOR and AMOS22, required to enable zero-shot and fine-tuning cross-dataset evaluation. Patient-level splits gather 804/160/107 patients for train/val/test. We additionally evaluate zero-shot and fine-tuned models on SegTHOR \cite{lambert2020segthor} and AMOS22 \cite{ji2022amos}. SegTHOR is a dataset of 40 CT volumes with annotations for esophagus, aorta and trachea, split at the patient level into train/val/test subsets. AMOS provides 200/100/200 CT scans for train/val/test, each with voxel-level annotations of 15 abdominal organs. We report results on the validation set, as the test set is not publicly available.


\noindent\textbf{Baseline architectures.} 
We integrate SegMate into three recent neural architectures: Efficient-NetV2-M 
\cite{tan2021efficientnetv2}, MambaOut-Tiny \cite{yu2024mambaout}, FastViT-T12 \cite{vasu2023fastvit}. We compare the original models (called vanilla) with the SegMate versions. These backbones are selected to cover different efficiency-focused architecture types and model sizes, including an efficient convolutional backbone, a gated CNN-style vision backbone, and a lightweight hybrid vision transformer. This selection allows us to evaluate whether SegMate behaves as a general architectural framework across different encoder families, or if its results are linked to a specific backbone. For each backbone, we compare the original model, denoted as vanilla, with the corresponding SegMate variant under the same training and evaluation protocol.

\begin{table*}[t]
\caption{Comparison with state-of-the-art methods across three datasets (3D volume-level Dice \%).
ZS: zero-shot transfer from TotalSegmentator; FT: fine-tuned.
$^\dagger$Models evaluated by PaR~\cite{huang2025par}.
$^\ddagger$Models evaluated by Bio2Vol~\cite{zhuang2025bio2vol}. Best per dataset is in \textbf{bold}.}
\label{tab:sota_comparison}
\vspace{-0.3cm}
\begin{center}
\begin{tabular}{|l|c|c|c|c|}
\hline
\textbf{Method} & \textbf{Type} & \textbf{ZS Dice} & \textbf{FT Dice} & \textbf{Train Set} \\
\hline
\multicolumn{5}{|l|}{\textit{TotalSegmentator}} \\
\hline
SegMate (EfficientNetV2-M) & 2.5D & -- & 93.51 & TotalSeg \\
nnU-Net~\cite{isensee2021nnu} & 2D & -- & \textbf{94.99} & TotalSeg \\
Standard U-Net & 2D & -- & 87.48 & TotalSeg \\
\hline
\multicolumn{5}{|l|}{\textit{SegTHOR}} \\
\hline
SegMate (EfficientNetV2-M) & 2.5D & \textbf{86.85} & 88.51 & TotalSeg (ZS) / SegTHOR (FT) \\
SegMate (MambaOut-T) & 2.5D & 86.46 & 86.67 & TotalSeg (ZS) / SegTHOR (FT) \\
SegMate (FastViT-T12) & 2.5D & 84.23 & 87.13 & TotalSeg (ZS) / SegTHOR (FT) \\
3D U-Net + PaR~\cite{huang2025par} & 3D & -- & \textbf{89.82}$^\dagger$ & SegTHOR \\
MultiResUNet~\cite{ibtehaz2020multiresunet} & 3D & -- & 88.53$^\dagger$ & SegTHOR \\
3D UX-Net~\cite{lee2023uxnet} & 3D & -- & 87.34$^\dagger$ & SegTHOR \\
Swin UNETR~\cite{hatamizadeh2022swin} & 3D & -- & 87.26$^\dagger$ & SegTHOR \\
SegResNet~\cite{myronenko2019segresnet} & 3D & -- & 86.99$^\dagger$ & SegTHOR \\
nnFormer~\cite{zhou2021nnformer} & 3D & -- & 86.65$^\dagger$ & SegTHOR \\
UNETR~\cite{hatamizadeh2022unetr} & 3D & -- & 84.03$^\dagger$ & SegTHOR \\
\hline
\multicolumn{5}{|l|}{\textit{AMOS22}} \\
\hline
SegMate (EfficientNetV2-M) & 2.5D & \textbf{89.35} & 91.29 & TotalSeg (ZS) / AMOS22 (FT) \\
SegMate (MambaOut-T) & 2.5D & 89.14 & 90.44 & TotalSeg (ZS) / AMOS22 (FT) \\
SegMate (FastViT-T12) & 2.5D & 88.87 & \textbf{91.52} & TotalSeg (ZS) / AMOS22 (FT) \\
Bio2Vol~\cite{zhuang2025bio2vol} & 3D & -- & 87.84$^\ddagger$ & AMOS22 \\
SwinUNETR~\cite{hatamizadeh2022swin} & 3D & -- & 84.37$^\ddagger$ & AMOS22 \\
SegMamba~\cite{xing2024segmamba} & 3D & -- & 84.26$^\ddagger$ & AMOS22 \\
nnFormer~\cite{zhou2021nnformer} & 3D & -- & 83.48$^\ddagger$ & AMOS22 \\
\hline
\end{tabular}
\end{center}
\end{table*}

\noindent\textbf{Pre-processing, hyperparameters, and evaluation settings.}
CT scans are processed into slices that are clipped to $[-1000, 2000]$ HU, normalized to $[0,1]$ and resized to $512\times512$ pixels. All models (including baselines) are optimized via AdamW ($2\cdot10^{-4}$ learning rate, $10^{-4}$ weight decay) with ``reduce on plateau'' scheduling and per-class loss weights to emphasize small organs and compensate for class imbalance. Training runs for 100 epochs on 3$\times$A100 GPUs with DDP, using a mini-batch size of 32. We reconstruct 3D volumes from predicted 2D slices for clinical relevance, and we report 3D volume-level Dice scores. Fine-tuning is performed for 25 epochs, with frozen encoders for the first 5 epochs, and a learning rate of $5\cdot10^{-6}$. All evaluations are performed on $1\times$GeForce RTX 5070 Ti GPU, measuring peak and average efficiency metrics on inference (GFLOPs, VRAM).


\subsection{Main results}
\noindent\textbf{Vanilla vs.~SegMate.}
In Table \ref{tab:unified_overview}, we report average Dice over all organs on TotalSegmentator (see third column). SegMate variants outperform vanilla variants for EfficientNetV2-M and MambaOut-T, with gains ranging from $+1.06\%$ (EfficientNetV2-M) to $+1.26\%$ (MambaOut-T), suggesting that our lightweight backbones benefit from the proposed architectural enhancements. Our SegMate (EfficientNetV2-M) variant reaches the highest Dice score ($93.51\%$) on TotalSegmentator.


\noindent\textbf{Cross-dataset generalization.}
In Table \ref{tab:unified_overview}, we also report zero-shot and fine-tuning results of the models that have been trained on TotalSegmentator. SegMate (EfficientNetV2-M) reaches $86.85\%$ zero-shot Dice, a $+1.45$ percentage points gain over the vanilla EfficientNetV2-M ($85.36\%$). 
After fine-tuning, SegMate (EfficientNetV2-M) achieves a Dice of $88.51\%$. 
On AMOS22, SegMate (EfficientNetV2-M) leads the zero-shot accuracy with $89.35\%$ average Dice across all five  organs. After fine-tuning, SegMate (FastViT-T12) achieves the best average Dice ($91.52\%$), demonstrating that the proposed framework can be competitive even with 120 MB memory usage. In general, SegMate reduces GFLOPs by up to $2.5\times$ and VRAM by up to $2.1\times$, while generally surpassing the vanilla models in terms of Dice.

\noindent\textbf{Comparison with state of the art.}
In Table \ref{tab:sota_comparison}, we compare SegMate against state-of-the-art methods.
On TotalSegmentator, SegMate (EfficientNetV2-M) reaches $93.51\%$ Dice at 295 MB peak VRAM, which is $40$-$50\times$ below the resources needed to run a 12-16GB state-of-the-art model such as Swin UNETR or 3D nnU-Net. On SegTHOR, our fine-tuned SegMate (EfficientNetV2-M) model ($88.51\%$) surpasses Swin UNETR and nnFormer, falling $1.31\%$ short of 3D U-Net+PaR \cite{huang2025par}. On AMOS22, all three zero-shot SegMate variants compare favorably with the literature-reported results, and fine-tuning further increases the margin. However, since the external baselines were trained and evaluated under different protocols, we treat these numbers as context rather than as evidence of a strict ranking.

\begin{table*}[t]
\caption{{3D volume-level Dice (\%) per organ on SegTHOR. Left: zero-shot transfer from TotalSegmentator. Right: fine-tuning on SegTHOR. Best score for each organ and each setup is in \textbf{bold}.}}
\label{tab:segthor_3d}
        \vspace{-0.3cm}

\begin{center}
{%
\begin{tabular}{|l|l|ccc c|ccc c|}
\hline
\multirow{2}{*}{\textbf{Backbone}} & \multirow{2}{*}{\textbf{Method}} & \multicolumn{4}{c|}{\textbf{Zero-shot}} & \multicolumn{4}{c|}{\textbf{Fine-tuned}} \\
\cline{3-10}
& & \textbf{Esoph.} & \textbf{Aorta} & \textbf{Trach.} & \textbf{Average} & \textbf{Esoph.} & \textbf{Aorta} & \textbf{Trach.} & \textbf{Average} \\
\hline
\multirow{2}{*}{EfficientNetV2-M}
  & SegMate & \textbf{82.43} & 86.75 & \textbf{91.37} & \textbf{86.85}$_{\pm2.79}$ & 80.86 & \textbf{94.10} & 90.58 & \textbf{88.51}$_{\pm3.32}$ \\
  & Vanilla & 80.54 & 85.87 & 89.79 & 85.40$_{\pm3.25}$ & \textbf{81.53} & 93.59 & 89.07 & 88.06$_{\pm2.90}$ \\
  \hline
\multirow{2}{*}{MambaOut-T}
  & SegMate & 81.76 & 86.83 & 90.81 & 86.46$_{\pm2.61}$ & 81.41 & 87.04 & \textbf{91.56} & 86.67$_{\pm3.03}$ \\
  & Vanilla & 79.64 & \textbf{86.88} & 90.82 & 85.78$_{\pm3.06}$ & 77.50 & 91.85 & 90.62 & 86.66$_{\pm4.35}$ \\
  \hline
\multirow{2}{*}{FastViT-T12}
  & SegMate & 79.12 & 84.53 & 89.05 & 84.23$_{\pm3.30}$ & 77.97 & 92.63 & 90.78 & 87.13$_{\pm3.94}$ \\
  & Vanilla & 75.14 & 86.34 & 88.09 & 83.19$_{\pm5.35}$ & 76.60 & 91.21 & 90.26 & 86.02$_{\pm4.76}$ \\
\hline
\end{tabular}%
}
\end{center}
\end{table*}

\begin{table*}[tb]
\caption{{3D volume-level Dice (\%) per organ on AMOS22 (validation set). Top: zero-shot transfer from TotalSegmentator. Bottom: fine-tuning on AMOS22. Best score for each organ and each setup is in \textbf{bold}.}}
\label{tab:amos22_3d}
        \vspace{-0.3cm}

\begin{center}
\begin{tabular}{|l|l|ccccc|c|}
\hline
\textbf{Backbone} & \textbf{Method} & \textbf{Esoph.} & \textbf{Liver} & \textbf{L.~Kid.} & \textbf{R.~Kid.} & \textbf{Aorta} & \textbf{Average} \\
\hline
\multicolumn{8}{|l|}{\textit{Zero-shot transfer}} \\
\hline
\multirow{2}{*}{EfficientNetV2-M}
  & SegMate & 78.51 & 91.10 & 92.70 & 92.36 & \textbf{92.09} & \textbf{89.35}$_{\pm5.44}$ \\
  & Vanilla & 78.35 & \textbf{91.31} & 92.46 & 92.30 & 92.03 & 89.29$_{\pm5.48}$ \\
 \hline
 
\multirow{2}{*}{MambaOut-T}
  & SegMate & \textbf{79.49} & 90.57 & 92.00 & 91.67 & 91.97 & 89.14$_{\pm4.85}$\\
  & Vanilla & 77.88 & 90.63 & 90.32 & 90.37 & 91.89 & 88.22$_{\pm5.20}$ \\
  \hline

\multirow{2}{*}{FastViT-T12}
  & SegMate & 76.59 & 90.78 & \textbf{93.23} & \textbf{92.58} & 91.16 & 88.87$_{\pm6.20}$ \\
  & Vanilla & 77.07 & 91.00 & 90.90 & 90.75 & 91.51 & 88.24$_{\pm5.59}$ \\
\hline
\multicolumn{8}{|l|}{\textit{Fine-tuned}} \\
\hline
\multirow{2}{*}{EfficientNetV2-M}
  & SegMate & 80.49 & \textbf{93.80} & 94.00 & 93.88 & \textbf{94.31} & 91.29$_{\pm5.40}$ \\
  & Vanilla & \textbf{81.48} & 93.63 & 93.60 & 93.33 & 94.08 & 91.22$_{\pm4.87}$\\
  \hline

\multirow{2}{*}{MambaOut-T}
  & SegMate & 79.29 & 93.20 & 93.25 & 92.71 & 93.75 & 90.44$_{\pm5.58}$\\
  & Vanilla & 79.96 & 92.63 & 93.18 & 92.64 & 92.96 & 90.28$_{\pm5.16}$ \\
  \hline

\multirow{2}{*}{FastViT-T12}
  & SegMate & 81.27 & 93.27 & \textbf{94.76} & \textbf{94.53} & 93.80 & \textbf{91.52}$_{\pm5.15}$ \\
  & Vanilla & 76.46 & 91.22 & 91.60 & 91.55 & 91.71 & 88.51$_{\pm6.02}$ \\
\hline
\end{tabular}
\end{center}
\end{table*}

\subsection{Per-organ and extended comparisons}
\noindent\textbf{Per-organ results}. 
In Tables \ref{tab:segthor_3d} and \ref{tab:amos22_3d}, we present the segmentation scores for each of the organs included in our cross-validation setups, comparing SegMate against its vanilla counterpart for each backbone. This comparison highlights the improvements brought by SegMate, especially for smaller organs, such as the esophagus and trachea.

\begin{table*}[t]
\caption{Cross-method comparison on SegTHOR (3D volume-level). SegMate uses a 28/7/5 split and evaluates the three organs shared by both settings (esophagus, aorta, and trachea); $^\dagger$PaR baselines~\cite{huang2025par} use a 32/4/4 split and run on all 4 organs (including heart).}
\label{tab:segthor_cross}
        \vspace{-0.3cm}

\begin{center}
\begin{tabular}{|l|c|c|c|l|}
\hline
\textbf{Method} & \textbf{Type} & \textbf{Dice (\%)} & \textbf{HD95 (mm)} & \textbf{Trained on} \\
\hline
\multicolumn{5}{|l|}{\textit{With PaR post-processing}} \\
\hline
3D U-Net + PaR~\cite{huang2025par} & 3D & \textbf{89.82}$^\dagger$ & \textbf{2.98} & SegTHOR \\
3D UX-Net + PaR~\cite{lee2023uxnet,huang2025par} & 3D & 87.77$^\dagger$ & 3.61 & SegTHOR \\

ResUNet + PaR~\cite{huang2025par} & 3D & 88.71$^\dagger$ & 3.14 & SegTHOR \\
MultiResUNet + PaR~\cite{ibtehaz2020multiresunet,huang2025par} & 3D & 89.10$^\dagger$ & 11.06 & SegTHOR \\
SegResNet + PaR~\cite{myronenko2019segresnet,huang2025par} & 3D & 89.60$^\dagger$ & 2.87 & SegTHOR \\

nnFormer + PaR~\cite{zhou2021nnformer,huang2025par} & 3D & 87.69$^\dagger$ & 3.51 & SegTHOR \\
Swin UNETR + PaR~\cite{hatamizadeh2022swin,huang2025par} & 3D & 87.42$^\dagger$ & 3.62 & SegTHOR \\
\hline
\multicolumn{5}{|l|}{\textit{Without PaR}} \\
\hline
SegMate (FT) & \textbf{2.5D} & 88.51$\pm$3.32 & 3.03$\pm$1.68 & TotalSeg (ZS) / SegTHOR (FT) \\
{SegMate (ZS)} & \textbf{2.5D} & 86.85$\pm$2.79 & 4.75$\pm$1.21 & {TotalSeg} (ZS) \\
\hline
3D U-Net~\cite{huang2025par} & 3D & 87.59$^\dagger$ & 4.32 & SegTHOR \\
3D UX-Net~\cite{lee2023uxnet,huang2025par} & 3D & 87.34$^\dagger$ & 4.69 & SegTHOR \\

ResUNet~\cite{huang2025par} & 3D & 88.26$^\dagger$ & 3.22 & SegTHOR \\
MultiResUNet~\cite{ibtehaz2020multiresunet,huang2025par} & 3D & 88.53$^\dagger$ & 26.75 & SegTHOR \\
SegResNet~\cite{myronenko2019segresnet,huang2025par} & 3D & 86.99$^\dagger$ & 3.38 & SegTHOR \\

nnFormer~\cite{zhou2021nnformer,huang2025par} & 3D & 86.65$^\dagger$ & 5.11 & SegTHOR \\
Swin UNETR~\cite{hatamizadeh2022swin,huang2025par} & 3D & 87.26$^\dagger$ & 3.87 & SegTHOR \\
UNETR~\cite{hatamizadeh2022unetr,huang2025par} & 3D & 84.03$^\dagger$ & 4.71 & SegTHOR \\

\hline
\end{tabular}
\end{center}
\end{table*}

\begin{table*}[t]
\caption{Per-organ comparison on AMOS22 validation set (5 overlapping organs, 3D volume-level). $^\ddagger$Baselines were retrained and evaluated by Bio2Vol~\cite{zhuang2025bio2vol} under their protocol.}
\label{tab:amos22_cross}
        \vspace{-0.3cm}

\begin{center}
\begin{tabular}{|l|c|c|c|c|c|c|}
\hline
\textbf{Method} & \textbf{Esoph.} & \textbf{Liver} & \textbf{L.~Kid.} & \textbf{R.~Kid.} & \textbf{Aorta} & \textbf{Average} \\
\hline
{SegMate (FT)} & \textbf{81.27} & 93.27 & \textbf{94.76} & \textbf{94.53} & 93.80 & \textbf{91.52}$_{\pm3.21}$ \\
SegMate (ZS) & 78.51 & 91.10 & 92.70 & 92.36 & 92.09 & 89.35$_{\pm4.59}$ \\
Bio2Vol~\cite{zhuang2025bio2vol} & 70.20 & \textbf{96.20} & 84.40 & 94.10 & \textbf{94.30} & 87.84$^\ddagger$ \\
SegMamba~\cite{xing2024segmamba,zhuang2025bio2vol} & 69.26 & 90.66 & 82.86 & 91.87 & 86.66 & 84.26$^\ddagger$ \\
SwinUNETR~\cite{hatamizadeh2022swin,zhuang2025bio2vol} & 68.55 & 91.28 & 82.40 & 91.27 & 88.36 & 84.37$^\ddagger$ \\
nnFormer~\cite{zhou2021nnformer,zhuang2025bio2vol} & 63.41 & 93.31 & 83.10 & 89.39 & 88.21 & 83.48$^\ddagger$ \\
UNETR~\cite{hatamizadeh2022unetr,zhuang2025bio2vol} & 65.27 & 89.51 & 72.81 & 90.05 & 78.62 & 79.25$^\ddagger$ \\
\hline
\end{tabular}%
\end{center}
\end{table*}
\noindent\textbf{Extended cross-dataset comparison.} 
In Tables \ref{tab:segthor_cross} and \ref{tab:amos22_cross}, we provide a more detailed comparison against state-of-the-art 3D baselines, adding reported HD95 scores on the SegTHOR dataset, and a per-organ breakdown with values identified in the literature for AMOS22. This comparison should be read together with the differences in training and evaluation methods noted in the captions. More precisely, the external baselines have been trained directly on the target datasets, and SegTHOR models are also trained on heart delineations.

In Table~\ref{tab:amos22_cross}, we present per-organ Dice on the five AMOS22 organs overlapping with our training set. Baseline values are taken from Zhuang et al.~\cite{zhuang2025bio2vol}, who retrained each architecture on AMOS22 under their own protocol.

\begin{table*}[t]
\caption{SegMate ablation for EfficientNetV2-M  applied on TotalSegmentator. Results are reported at 2D slice level (organ-averaged). Values after vanilla baseline show absolute percentage point change w.r.t~the baseline. Highest gain is in \textbf{bold}.}
\label{tab:arch_ablation}
        \vspace{-0.3cm}

\begin{center}
\begin{tabular}{|l|c|c|c|c|c|c|}
\hline
\textbf{Measure} & \textbf{Vanilla} & \textbf{+Asym. dec.} & \textbf{+CBAM} & \textbf{+SE} & \textbf{+2.5D} & \textbf{+Slice Pos.} \\
\hline
Dice (\%) $\uparrow$ & 92.45 & +0.45 & +0.65 & +0.81 & +0.95 & \textbf{+1.06} \\
\hline
HD95 (mm) $\downarrow$ & 4.97 & $-$0.62 & $-$0.61 & $-$0.80 & $-$0.81 & \textbf{$-$0.84} \\
\hline
VRAM (MB) $\downarrow$ & 373.9 & \textbf{$-$85.5} & \textbf{$-$85.5} & $-$85.4 & $-$85.4 & \textbf{$-$85.5} \\
\hline
GFLOPs/slice $\downarrow$ & 37.25 & \textbf{$-$5.79} & $-$5.78 & $-$5.78 & $-$5.78 & $-$5.78 \\
\hline
\end{tabular}
\end{center}
\end{table*}

\begin{figure*}[!t]
    \begin{center}
    \includegraphics[width=0.55\linewidth]{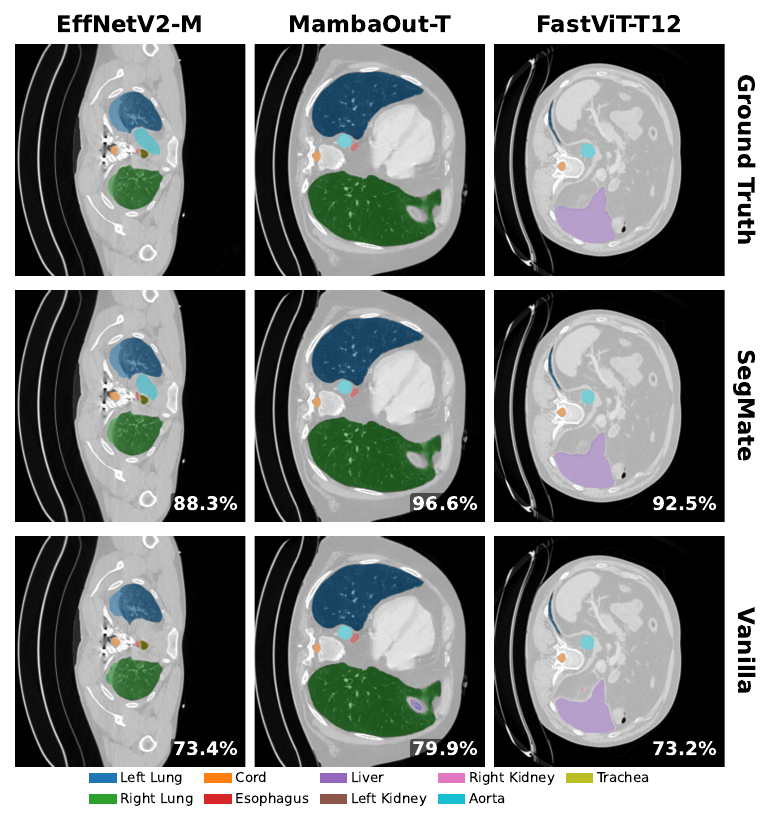}
    \end{center}
        \vspace{-0.5cm}

    \caption{Qualitative comparison on TotalSegmentator test slices. Vanilla EfficientNetV2-M fails to segment the aorta entirely. Vanilla MambaOut-Tiny incorrectly predicted the liver delineations. Vanilla FastViT-T12 wrongly identifies the right kidney in the slice. Best viewed in color.}
    \label{fig:segmentations1}
\end{figure*}

\begin{figure*}[!t]
    \begin{center}
    \includegraphics[width=0.55\linewidth]{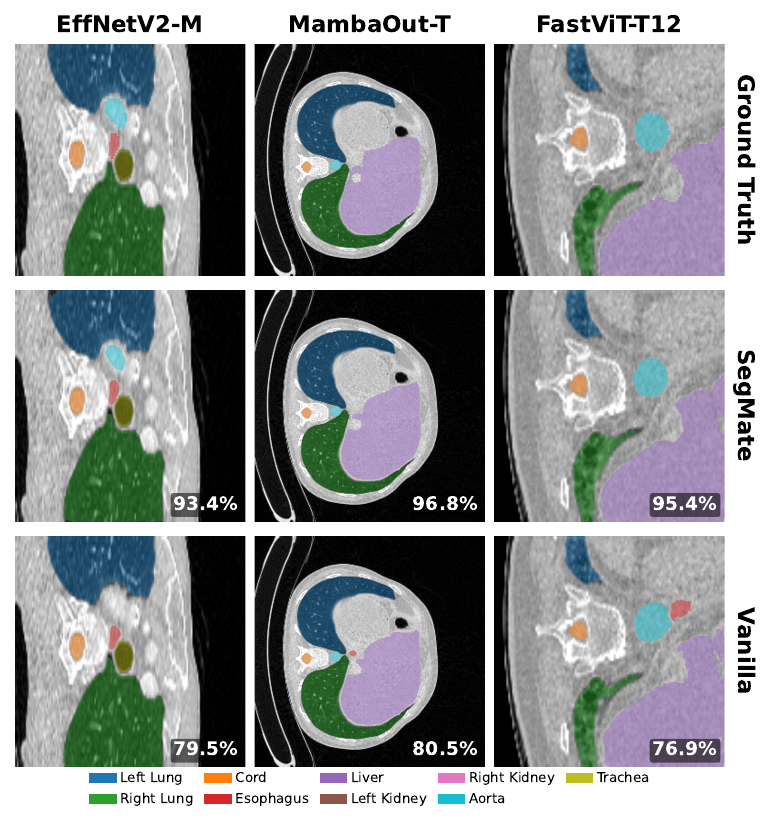}
    \end{center}
    \vspace{-0.5cm}
    \caption{Qualitative comparison on TotalSegmentator test slices.
    SegMate closely matches the ground truth across all three backbones.
    The vanilla EfficientNetV2-M misses the aorta, while vanilla MambaOut-T
    and FastViT-T12 produce false-positive esophagus segmentations. Best viewed in color.}
    \label{fig:segmentations2}
\end{figure*}

\begin{table*}[tb]
\caption{{Per-volume GFLOPs for a $512\times512\times162$ CT volume. 2.5D models process 162 sequential slices with stride 1 ($66\%$ overlap). 3D models use sliding-window inference over $96^3$ patches with stride~48 ($50\%$ overlap), which amounts to ${\approx}$200 patches. 3D figures are derived from the per-patch costs reported by EffiDec3D~\cite{rahman2025effidec3d} and EfficientMedNeXt~\cite{rahman2025efficientmednext}. We mark the lowest cost within each group in \textbf{bold}.}}
\label{tab:gflops_comparison}
        \vspace{-0.3cm}

\begin{center}
\begin{tabular}{|l|l|c|c|c|c|}
\hline
\textbf{Method} & \textbf{Paradigm} & \textbf{Params (M)} & \textbf{GFLOPs/unit} & \textbf{Units/vol} & \textbf{GFLOPs/vol} \\
\hline
\multicolumn{6}{|l|}{\textit{SegMate (2.5D, $512\times512$ slices $\times$ 162)}} \\
\hline
SegMate (EfficientNetV2-M)    & 2.5D & 58.3 & 31.46 & 162 & 5,097 \\
SegMate (MambaOut-Tiny) & 2.5D & 29.4 & 27.06 & 162 & 4,384 \\
SegMate (FastViT-T12)   & 2.5D & 12.4 & 9.11  & 162 & \textbf{1,476} \\
\hline
\multicolumn{6}{|l|}{\textit{Efficient 3D methods ($96^3$ patches $\times$ 200)}} \\
\hline
EfficientMedNeXt-T~\cite{rahman2025efficientmednext}      & 3D & 0.43 & 7.09  & 200 & \textbf{1,418} \\
EfficientMedNeXt-L~\cite{rahman2025efficientmednext}      & 3D & 2.19 & 57.68 & 200 & 11,536 \\
3D UX-Net + EffiDec3D~\cite{rahman2025effidec3d}    & 3D & 3.16 & 51.47 & 200 & 10,294 \\
SwinUNETR + EffiDec3D~\cite{rahman2025effidec3d}    & 3D & 18.2 & 89.27 & 200 & 17,854 \\
SwinUNETRv2 + EffiDec3D~\cite{rahman2025effidec3d}  & 3D & 18.2 & 88.36 & 200 & 17,672 \\
MedNeXt-M K3 + EffiDec3D~\cite{rahman2025effidec3d} & 3D & 5.77 & 49.73 & 200 & 9,946 \\
\hline
\multicolumn{6}{|l|}{\textit{Standard 3D baselines ($96^3$ patches $\times$ 200)}} \\
\hline
3D UX-Net    & 3D & 53.0 & 631.97 & 200 & 126,394 \\
SwinUNETR    & 3D & 62.0 & 641.69 & 200 & 128,338 \\
nnU-Net      & 3D & 31.8 & 417.96 & 200 & 83,592 \\
MedNeXt-M K3 & 3D & 17.6 & 110.65 & 200 & 22,130 \\
UNETR        & 3D & 92.6 & 82.60  & 200 & \textbf{16,520} \\
\hline
\end{tabular}
\end{center}
\end{table*}

\subsection{Ablation and efficiency analysis}
\subsubsection{Ablation study}
In Table~\ref{tab:arch_ablation}, we present an ablation study where we progressively add each SegMate component to the vanilla baseline (EfficientNetV2-M). Our results show that each step improves Dice and HD95, while the asymmetric decoder alone cuts VRAM and GFLOPs by a large extent. 

To quantify the performance boost driven by recovering volumetric context via FiLM, we evaluate SegMate with and without slice-positioning conditioning. The results show that for the EfficientNetV2-M backbone, removing FiLM lowers the zero-shot SegTHOR Dice from $86.85\%$ to $84.86\%$ ($-1.99$), and the TotalSegmentator Dice by $0.45$ percentage points. On SegTHOR, the largest drop is registered for the esophagus ($-2.62$), whose appearance varies more strongly along the Z-axis. This indicates that the global context in regard to the slice position along the vertical axis of the CT volume, which is lost in the 2.5D to 2D reduction, is significantly recovered.


\subsubsection{Efficiency analysis}
In Figure \ref{fig:tradeoff}, we showcase the efficiency-accuracy trade-off measured on the TotalSegmentator dataset. All three SegMate variants register VRAM below 300 MB and Dice scores between $92.2\%$ and $93.6\%$. 
SegMate variants based on the FastViT-T12 encoder reach the extreme efficiency end (120 MB VRAM with 1,476 GFLOPs), while maintaining a competitive Dice of $92.25\%$.
To quantify total compute per patient, we measure per-slice GFLOPs and multiply by the average number of axial slices (162 slices on average in our test set). SegMate (FastViT) segments a full volume in 1,476 GFLOPs, while the EfficientNetV2-M variant requires 5,097 GFLOPs. As reference, recent efficient 3D methods such as EfficientMedNeXt-Tiny~\cite{rahman2025efficientmednext} and EffiDec3D~\cite{rahman2025effidec3d} report low per-patch costs (7-51 GFLOPs), but sliding-window inference over the same volume size still totals ${\sim}$1,500-10,000 GFLOPs. 

\begin{table*}[!t]
\caption{Inference efficiency in terms of various dimensions measured on RTX 5070 Ti.}
\label{tab:efficiency}
        \vspace{-0.1cm}

\centering
\begin{tabular}{|l|c|c|c|c|c|}
\hline
\multirow{2}{*}{\textbf{Backbone}} & \textbf{Params} & \textbf{GFLOPs} & \textbf{Speed} & \textbf{Throughput} & \textbf{VRAM} \\
         & \textbf{(M)}    & \textbf{/vol}   & \textbf{(ms)} & \textbf{(slices/s)}& \textbf{(MB)} \\
\hline
EffNetV2-M & 58.3 & 5097 & 28.6 & 34.9 & 295 \\
MambaOut-T & 29.4 & 4384 & 21.6 & 46.3 & 208 \\
FastViT-T12& 12.4 & 1476 & 19.9 & 50.3 & 120 \\
\hline
\end{tabular}
\end{table*}

In Table \ref{tab:gflops_comparison}, we show an explicit comparison of per-volume efficiency between SegMate, standard 3D baselines and state-of-the-art efficient architectures. While the SegMate variant (FastViT-T12) with the lowest GFLOPs per volume challenges the EfficientMedNeXt-T model, even our heaviest backbone stays below every EffiDec3D variant, and roughly 2.3$\times$ below EfficientMedNeXt-L. 

In Table \ref{tab:efficiency}, we report the measured inference cost for each SegMate backbone reported on a setup with one RTX 5070 Ti: parameter count, per-volume GFLOPs, average slice latency, throughput and peak VRAM. The report highlights that  the heaviest variant of SegMate (EfficientNetV2-M) runs at roughly 35 slices per second requiring 295 MB VRAM, while the most compact variant of SegMate (FastViT-T12) reaches 50 slices/s and 120 MB, allowing the models to run on-premise on modest hardware available in most clinical settings. The measured inference time is not exactly proportional to the GFLOPs, as it depends on the memory-access mechanism, so the latency in Table \ref{tab:efficiency} should be read as indicative, not as hardware-independent.

For archive-scale workloads, slice independence enables batched inference. Measured on pre-loaded $512{\times}512$ slices with mixed precision (thus excluding disk I/O), SegMate (FastViT-T12) sustains 463 slices/s at a mini-batch size of 8, i.e.~over 10,000 volumes per hour within 341 MB of peak VRAM, while EfficientNetV2-M reaches ${\sim}$7,000 volumes per hour under 750 MB. Sampling the GPU power draw via \texttt{nvidia-smi} (24 W idle, 88-267 W under load) yields 74-120 J per volume at the most efficient operating points, meaning that segmenting 1,000 CT volumes costs only 21-33 Wh of energy. At these rates, re-segmenting the entire TotalSegmentator corpus (1,228 volumes) requires 7-11 minutes of GPU compute on a single RTX 5070 Ti, making storage bandwidth the practical bottleneck in population-scale processing, rather than the ML model.

\subsection{Qualitative results}
In Figures~\ref{fig:segmentations1} and~\ref{fig:segmentations2}, we show OAR segmentation results on test slices from TotalSegmentator, comparing SegMate against the vanilla baseline across all three backbones. The examples illustrate cases where the vanilla model misses or mislabels an organ that SegMate segments correctly.
In general, the differences are visible both around small structures, such as the esophagus, trachea and aorta, and also in false-positive regions where the vanilla models wrongly assign labels to anatomically implausible areas. These examples support the quantitative trends observed in the per-organ results, where SegMate improves the delineation of challenging structures, while reducing large local errors. 

\section{Discussion}
Our work shows results over three datasets for SegMate variations based on three backbones. Across all trials, the architecture sits on the efficiency-accuracy frontier, proving itself as an enabler for efficient encoders to drive the accuracy to near state-of-the-art results. The backbones included in this study cover a practical range, from EfficientNetV2-M for the highest accuracy to FastViT-T12 for the lowest memory requirements and fastest inference. All three models run under 300 MB VRAM, a significant gap from the heavy setups needed by 3D convolutional and transformer architectures that produce similar or marginally higher accuracy levels.

The results of the ablation studies and per-organ reports point to a consistent pattern. The components that raise the performance of the vanilla baselines are the ones that restore the information a traditional model based on 2D convolutions would miss. Slice-position conditioning (FiLM) and the dual attention mechanism (SE and CBAM) improve delineations especially for small/thin structures, such as the esophagus, trachea, and spinal cord. Moreover, these components also drive the performance in zero-shot inference on external datasets, generalizing the interpretation of the CT slices. This confirms the intuition that the regular 2.5D models perform worse in areas where 3D context is significant, and that a light positional signal can recover part of that information.

As shown in Tables \ref{tab:segthor_3d}, \ref{tab:segthor_cross} and \ref{tab:amos22_cross}, we report the standard deviation of average Dice across the test patients, with each score computed as the mean of three runs to reduce the effect of lucky draws and fairly compare models.

\section{Limitations} 

Our evaluation covers a 9-organ subset dictated by the label overlap required for cross-dataset transfer; since only the three $1{\times}1$ output heads scale with the number of classes, the architecture itself does not restrict the organ count, but validating this on the full TotalSegmentator label set remains future work. The slice position $z_{\text{norm}}$ is defined per scan, so partially acquired volumes can map the same anatomy to different positions, which may affect robustness on cropped fields of view. Finally, the external baselines in Tables \ref{tab:sota_comparison}, \ref{tab:segthor_cross} and \ref{tab:amos22_cross} follow their original authors' protocols and organ sets, so those comparisons are indicative rather than strictly controlled, and AMOS22 results are reported on the validation split, as the test labels are not public.


\section{Conclusion}

In this paper, we introduced SegMate, a framework that integrates multiple architectural design changes that target efficiency boosting and effectiveness preservation for organ segmentation in medical images. We presented empirical evidence to demonstrate that SegMate achieves the desired goals across distinct neural architectures, datasets and evaluation setups. Furthermore, we showed the effect of introducing the proposed components via an ablation study, confirming that each component plays a key role in either reducing computation or boosting performance. 

In future work, we aim to integrate SegMate into additional models and construct a more comprehensive repository of pre-trained readily available models at \url{https://github.com/andreibunea99/SegMate}.

\section{Acknowledgments} 

This work was supported by a grant of the Ministry of Research, Innovation and Digitization, CCCDI - UEFISCDI, project number PN-IV-P7-7.1-PED-2024-1856, within PNCDI IV.
%
%
%
\bibliographystyle{IEEEtran}
\bibliography{ref}

\end{document}